%% file: main.tex
\documentclass[10pt,twocolumn,letterpaper]{article}

\usepackage{iccv}
\usepackage{times}
\usepackage{epsfig}
\usepackage{graphicx}
\usepackage{xcolor}
\usepackage{amsmath}
\usepackage{amssymb}
\usepackage{subcaption}
\usepackage{xspace}

\usepackage[pagebackref=true,breaklinks=true,letterpaper=true,colorlinks,
  citecolor=citecolor,bookmarks=false]{hyperref}
\definecolor{citecolor}{RGB}{34,139,34}

\def\iccvPaperID{}\iccvfinalcopy

\ificcvfinal\fi

\newcommand{\ourparagraph}[1]{\vspace{1.5mm}\noindent\textbf{#1}}
\newcommand{\tablestyle}[2]{\setlength{\tabcolsep}{#1}\renewcommand{\arraystretch}{#2}\centering\footnotesize}
\newlength\savewidth\newcommand\shline{\noalign{\global\savewidth\arrayrulewidth
  \global\arrayrulewidth 1pt}\hline\noalign{\global\arrayrulewidth\savewidth}}

\newcommand{\app}{\raise.17ex\hbox{$\scriptstyle\sim$}}

\DeclareMathOperator*{\argmin}{arg\,min}

\newcommand{\method}{RHO\xspace}
\newcommand{\dataset}{MOW\xspace}

\newcommand{\customfootnotetext}[2]{{% Group to localize change to footnote
  \renewcommand{\thefootnote}{#1}% Update footnote counter representation
  \footnotetext[0]{#2}}}% Print footnote text

\begin{document}
\title{Reconstructing Hand-Object Interactions in the Wild}
\author{%
 Zhe Cao$^{*}$ \quad Ilija Radosavovic$^{*}$ \quad Angjoo Kanazawa \quad Jitendra Malik\\[2mm]
 University of California, Berkeley}

\twocolumn[{%
\renewcommand\twocolumn[1][]{#1}%
\maketitle
\begin{center}
    \centerline{\includegraphics[width=1.0\linewidth]{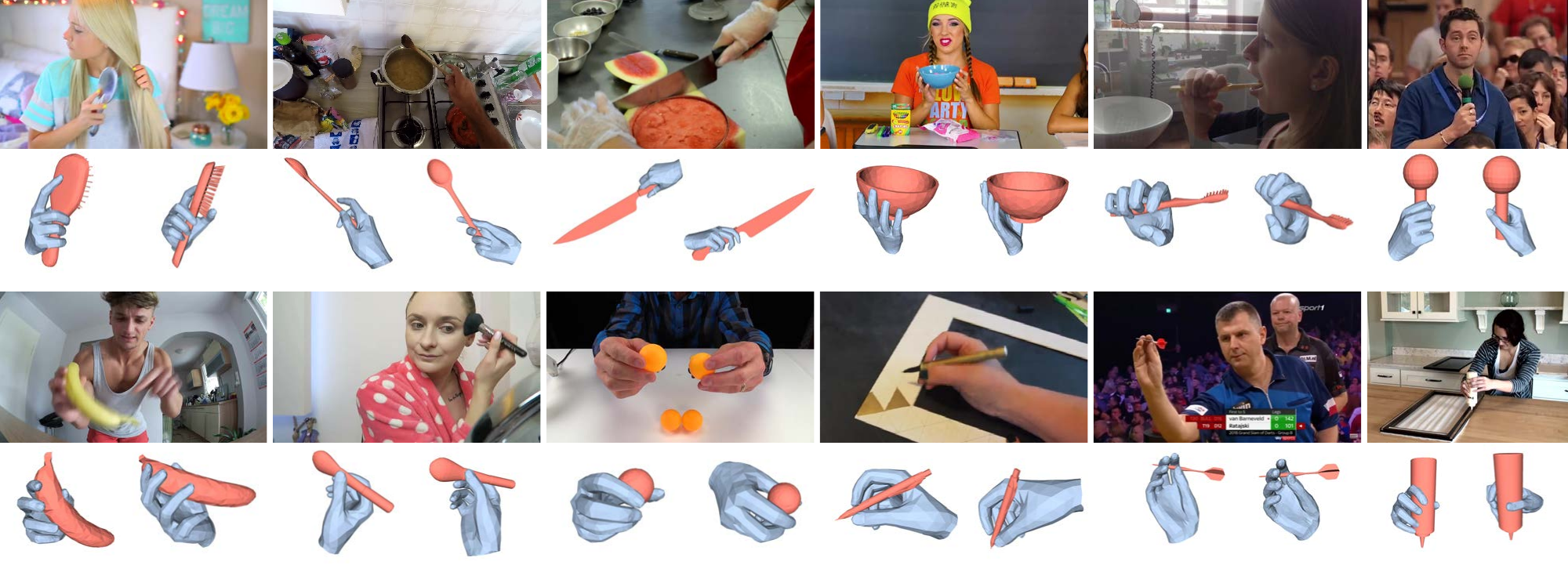}}
    \vspace{-2mm}
    \captionof{figure}{\small \textbf{Reconstructions in the wild.} For each row, we show the input image (top), the reconstructed hand and object in two different viewpoints (bottom). Our method can achieve compelling results for a variety of object categories, grasp types, and interaction scenarios. }\label{fig:teaser}
    %\vspace{-1mm}
\end{center}%
}]

%%%%%%%%%%%%%%%%%%%%%%%%%%%%%%%%%%%%%%%%%%%%%%%%%%%%%%%%%%%%%%%%%%%%%%%%%%%%%%%%%%%%%%%%%%%%%%%%%%%
\begin{abstract}
\vspace{-1.0mm}
We study the problem of understanding hand-object interactions from 2D images in the wild. This requires reconstructing both the hand and the object in 3D, which is challenging because of the mutual occlusion between the hand and the object. In this paper we make two main contributions: (1) a novel reconstruction technique, RHO (Reconstructing Hands and Objects), which reconstructs 3D models of both the hand and the object leveraging the 2D image cues and 3D contact priors; (2) a dataset MOW (Manipulating Objects in the Wild) of 500 examples of hand-object interaction images that have been "3Dfied" with the help of the RHO technique. Overall our dataset contains 121 distinct object categories, with a much greater diversity of manipulation actions, than in previous datasets.
\vspace{-2mm}
\end{abstract}

\customfootnotetext{*}{Equal contribution. Dataset available at this \href{https://people.eecs.berkeley.edu/~zhecao/rhoi}{project page}.}

%%%%%%%%%%%%%%%%%%%%%%%%%%%%%%%%%%%%%%%%%%%%%%%%%%%%%%%%%%%%%%%%%%%%%%%%%%%%%%%%%%%%%%%%%%%%%%%%%%%
\section{Introduction}

Our hands are the primary way we interact with objects in the world. In turn, we designed our world with hands in mind. Therefore, understanding hand-object interactions is an important ingredient for building agents that perceive and act in the real world. For example, it can allow them to learn object affordances~\cite{Gibson1977}, infer human intents~\cite{Meltzoff1995}, and learn manipulation skills from humans~\cite{Radosavovic2021, Mandikal2020, 2018-TOG-SFV}.

What does it mean to understand hand-object interactions? We argue that fully capturing the richness of hand-object interactions requires 3D understanding. In general, understanding 3D from a single RGB image is an under-constrained problem. In the case of hand-object interactions, the problem is even more challenging due to heavy occlusions that occur during object manipulation, a wide range of small daily objects that are not even present in labeled recognition datasets, and fine-grained interactions with complex contacts that are difficult to model.

%##################################################################################################
\begin{figure}[t]\centering
\includegraphics[width=0.9\linewidth]{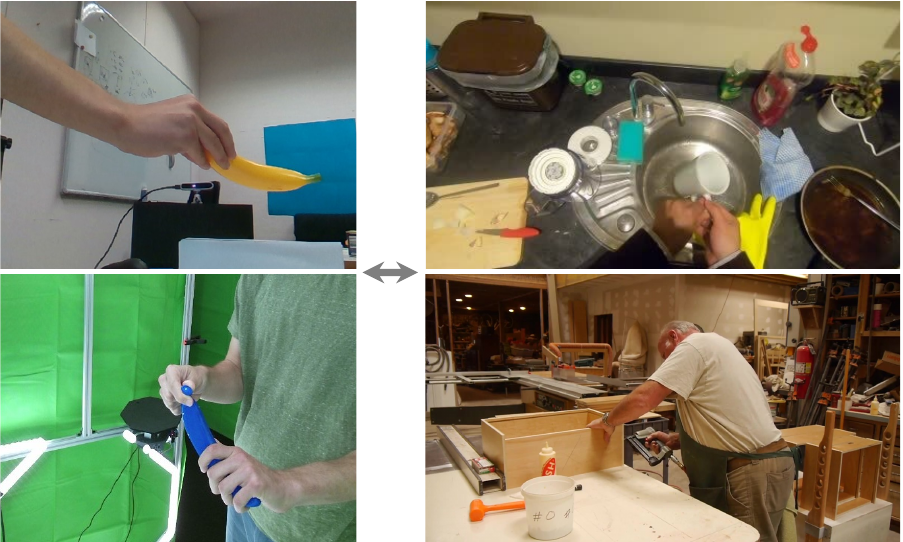}\vspace{-0mm}
\caption{\textbf{Images from existing hand-object dataset.} The reality gap between the existing in-the-lab datasets with 3D annotations (left) and in-the-wild images (right) is large.}
\label{fig:data}\vspace{-3mm}
\end{figure}
%##################################################################################################

Overall, our community has made substantial progress toward this goal. However, due to the difficulty in obtaining 3D annotations in the wild, the data collection efforts have focused mainly on in-the-lab setting~\cite{hampali2020honnotate, zimmermann2019freihand, hernando2018cvpr, Brahmbhatt_2020_ECCV, GRAB:2020}. As shown in Figure~\ref{fig:data}, there is a large reality gap between the existing in-the-lab settings and the richness of environments and interactions found in images in the wild. Indeed, as shown in Table~\ref{tab:datasets}, existing datasets have a limited number of participants and objects.

In this paper, we make two main contributions: (1) we develop a new technique for reconstructing 3D hands and objects from single images \emph{in-the-wild}, called RHO for Reconstructing Hands and Objects and (2) we use this technique in conjunction with human intervention to create a new 3D dataset of humans Manipulating Objects in-the-Wild, that we call \emph{\dataset}.

Specifically, \emph{\method} is a new optimization-based method for reconstructing hand-object interactions in the wild. The core idea is to leverage 2D image cues and 3D contact priors to provide constraints. \method consists of four steps: hand pose estimation using 2D hand keypoints, object pose estimation using 2D object mask and depth via differentiable rendering, joint optimization for hand-object configuration in 3D, and pose refinement using 3D contact priors.

A nice property of our method is the ability to recover a wide variety of objects in the wild---an order of magnitude more than previous work in areas of human-object or hand-object reconstruction. This required several innovations. First, an observation that segmentation masks for unknown object categories can be obtained using available recognition models. Second, a scalable data-driven way to enforce contact priors that is learned from 3D MoCap data recorded in the lab with instruments.

We compare \method to existing approaches quantitatively in the lab settings where ground truth annotations are available and qualitatively on in-the-wild images. We find that \method performs better or on par with the state-of-the-art method on in-the-lab datasets. Moreover, we show that the existing approaches struggle on challenging in-the-wild images reinforcing the need for the dataset we collect.

We use our method as a tool for data annotation and involve human intervention for two reasons. First, to find and prepare the appropriate 3D model for the object being manipulated in the image. Second, to ensure high quality annotations by verifying and adjusting the results of our method in an iterative fashion. Using this procedure, we annotate 500 images from the EPIC Kitchens~\cite{Damen2018} and the 100 Days of Hands~\cite{Shan2020} datasets. These depict a rich diversity of manipulation actions, which we augment with newly collected 3D object models from 121 object categories, 3D object poses, and 3D hand poses.

%##################################################################################################
\begin{table}[t]\centering
\resizebox{\columnwidth}{!}{\tablestyle{8pt}{1.05}
\begin{tabular}{@{}l|ccc|c@{}}
                & HO3D~\cite{hampali2020honnotate}    & CP~\cite{Brahmbhatt_2020_ECCV}     & GRAB~\cite{GRAB:2020}   & Ours         \\\shline
 setting        & lab             & lab                     & lab        & wild         \\
 data type      & video           & video                   & MoCap      & image        \\
 particip.       &         10      &  50                     &   10       & 450          \\
 objects        &         10      &  25                     &   51       & 121          
\end{tabular}}\vspace{1mm}
\caption{\textbf{Existing 3D hand-object datasets.} Our dataset contains \emph{in-the-wild} images, as shown in Figure~\ref{fig:data} right, and a large number of different participants and objects.}
\label{tab:datasets}\vspace{-3mm}
\end{table}
%##################################################################################################

The resulting dataset can be used in a number of ways. We highlight two potential use-cases next. First, the dataset can be used to evaluate hand-object reconstruction methods on challenging in-the-wild data. Second, it can be used to learn more about human manipulation from images in the wild. For example, we present initial analysis of our data and observe interesting trends (Figure~\ref{fig:dist_emb}).

In summary, our key contributions are: (1) We present a novel optimization-based procedure, \method, that is able to reconstruct hand-object interactions in the wild across diverse object categories; (2) We show quantitative and qualitative improvements over existing methods, especially on in-the-wild setting; (3) We contribute a new 3D dataset, \dataset, of 500 images in the wild, spanning 121 object categories with annotation of instance category, 3D object models, 3D hand pose, and object pose annotation.

We encourage the readers to see the \href{https://people.eecs.berkeley.edu/~zhecao/rhoi}{project page} for additional materials and data download instructions.

%%%%%%%%%%%%%%%%%%%%%%%%%%%%%%%%%%%%%%%%%%%%%%%%%%%%%%%%%%%%%%%%%%%%%%%%%%%%%%%%%%%%%%%%%%%%%%%%%%%
\input{Section/related_work}

%%%%%%%%%%%%%%%%%%%%%%%%%%%%%%%%%%%%%%%%%%%%%%%%%%%%%%%%%%%%%%%%%%%%%%%%%%%%%%%%%%%%%%%%%%%%%%%%%%%
\section{Method}\label{sec:method}

\begin{figure}[t]\centering
\includegraphics[width=1\linewidth]{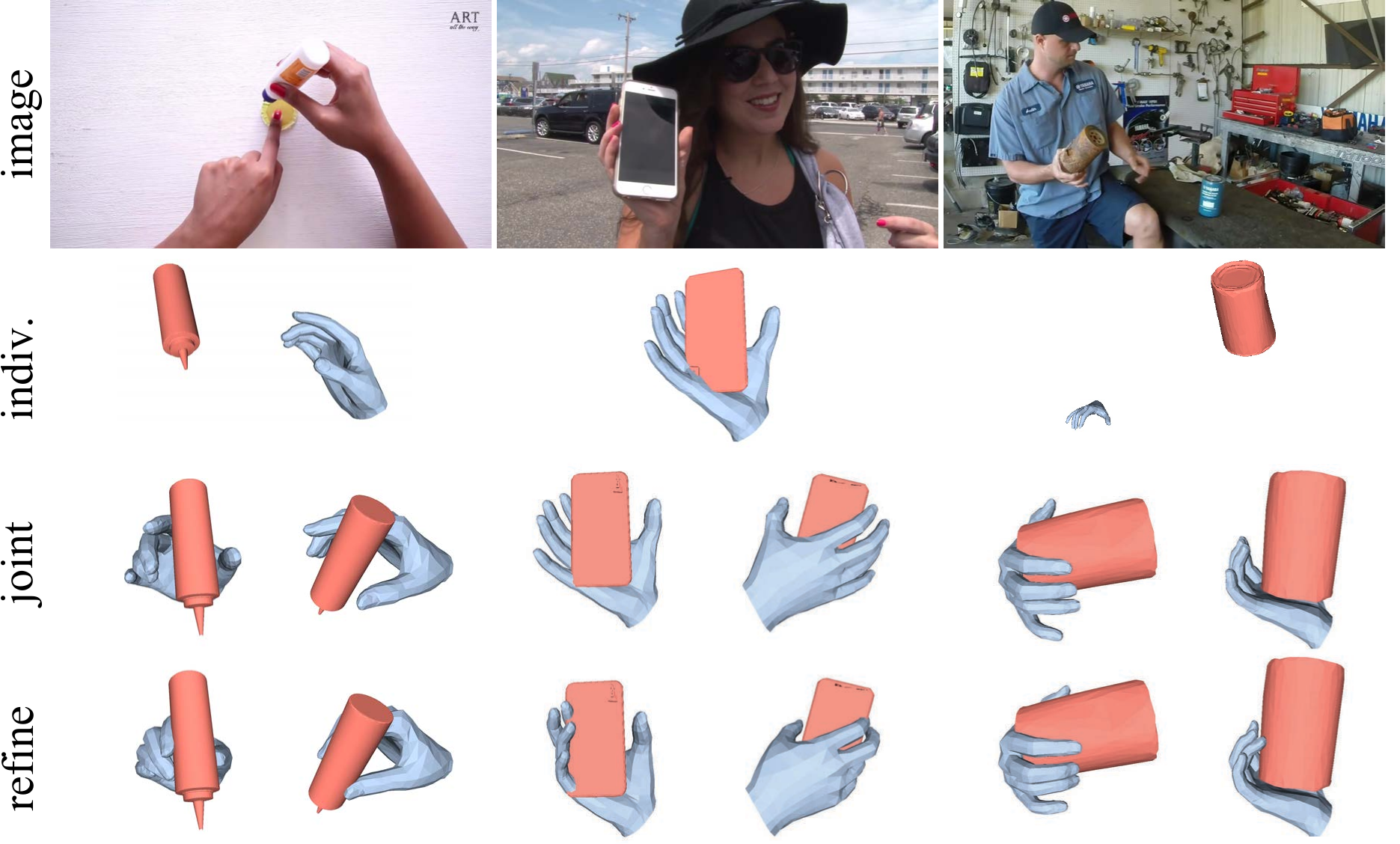}\vspace{-0mm}
\caption{\textbf{Intermediate results.} \emph{Top row:} input images. \emph{2nd row:} results from individually optimizing hand and object. \emph{3rd row:} results from joint optimization (two viewpoints per example). \emph{Bottom row:} results after refinement.}
\label{fig:qual_steps}\vspace{-2mm}
\end{figure}

We first describe our method for reconstructing hand-object interactions in the wild, called \emph{\method}. As shown in Figure~\ref{fig:pipeline}, it involves 4 steps: estimating the hand pose, the object pose, their 3D configuration jointly, and finally refining the pose using 3D contact priors. Intermediate results from each step are shown in Figure~\ref{fig:qual_steps}. We describe each step next. We note that while \method can be applied to multiple hands and objects, we assume a single pair for brevity.
We will evaluate \method, discuss how we curate our new dataset \dataset, and present  analysis in the following sections.

\subsection{Hand Pose Estimation}

The first step of \method involves hand pose estimation (Figure~\ref{fig:pipeline}a). Given an input image, we aim to reconstruct the full 3D hand mesh. We use a learning-based method to obtain the initial result and further improve the estimation by fitting it to 2D hand keypoints.

In particular, we represent the hand using a parametric model defined by MANO~\cite{romero2017embodied}: $\boldsymbol{V}_h =  H(\boldsymbol{\theta}, \boldsymbol{\beta})$, where $\boldsymbol{\theta} \in \mathbb{R}^{3 \times 15}$ and $\boldsymbol{\beta} \in \mathbb{R}^{10}$ are the pose and shape parameters, respectively. Taking a single RGB image as input, we use FrankMocap~\cite{rong2020frankmocap} to estimate the weak-perspective camera model ${\Pi}_h = (t_x, t_y, s_h)$, and initial 3D hand parameters $\boldsymbol{\theta}$ and $\boldsymbol{\beta}$. We further optimize the hand pose by fitting to 2D hand keypoints obtained from~\cite{8765346, simon2017hand}.

The hand pose optimization objective is to minimize the difference between 2D keypoints detection and 2D projection of 3D hand keypoints:
\begin{equation}
\small
\boldsymbol{\theta}^{*}, \boldsymbol{\beta}^{*} = \arg\min_{\boldsymbol{\theta}, \boldsymbol{\beta}}\, L_{joints} + L_{reg},
\label{eq::pose}
\end{equation}
consisting of a 2D keypoints distance term $L_{joints}$ and a regularization term $L_{reg}$ for hand shape $\boldsymbol{\beta}$.

We convert the weak-perspective to perspective camera by assuming a fixed focal length $f$. The depth of the hand is approximated by the focal length divided by the camera scale  $s_h$. We obtain the final hand vertices by:
\begin{equation}
\small
	\boldsymbol{V}^{*}_h =  H(\boldsymbol{\theta}^{*}, \boldsymbol{\beta}^*) + [t_x, t_y, f/s_h], \\
\end{equation}

%##################################################################################################
%##################################################################################################

\subsection{Object Pose Estimation}

In the next step of our method, Figure~\ref{fig:pipeline}b, we recover the object pose using an \emph{analysis-by-synthesis} approach. Given an input image and 3D model, we want to optimize the object scale $s \in \mathbb{R}$, 3D rotation $\boldsymbol{R} \in SO(3)$, and translation $\boldsymbol{T} \in \mathbb{R}^{3}$. We use a differentiable renderer~\cite{kato2018renderer} to render 3D model into 2D mask and depth maps. By comparing the rendered mask/depth with the targets, we compute the gradients to update the object parameters.

\ourparagraph{Object mask estimation.} How can we obtain good objects masks for diverse objects in images in the wild? Modern 2D recognition models trained on large labeled datasets can provide reasonable predictions on real-world data. However, in our case, we require instance masks for a range of object categories that are not even present in the available labeled datasets (\eg, spatula, pliers, mic, \etc). Thus, we cannot expect the available models to recognize the objects correctly in our setting. We observe that even if the predicted categories are incorrect, the instance masks are still quite reasonable for a variety of objects. For example, the models do not know what a spatula is called but are still able to segment it (see Figure~\ref{fig:mask_cat} for examples).

With this observation, we use available recognition models to estimate instance mask ignoring the category information. Specifically, we use PointRend model~\cite{kirillov2020pointrend} trained on COCO~\cite{Lin2014}. For all object instances predictions in the image, we decide the instance that the hand is interacting with by running a hand detector~\cite{Shan2020}. Namely, we select the instance with highest IoU with the detected hand box. This automatic way allows us obtain instance masks for more than 100 daily object categories as shown in Section~\ref{sec:dataset:analysis}.

\ourparagraph{Mask loss.} Given the estimated object mask, we optimize the object pose via differentiable rendering. In particular, we define the object mask loss as the L1 difference between the rendered and the estimated object masks: 
\begin{equation}
\small
	L_{mask} = \lVert \textrm{NR}_m (s, \boldsymbol{R}, \boldsymbol{T}) - \boldsymbol{M} \rVert,\\
\end{equation}
\noindent
where $\textrm{NR}(\cdot)$ denotes the differentiable rendering operation which renders the 3D mesh into the 2D mask.

\ourparagraph{Depth loss.} While the 2D mask loss is sufficient in some cases, it does not capture geometry information and can be ambiguous---multiple object poses can lead to similar 2D masks (see Figure~\ref{fig:qual_depth} for examples). To overcome this, we employ a new loss term which fits 3D model to the depth map $\boldsymbol{D}$ estimated using~\cite{Ranftl2020}:
\begin{equation}
\small
	L_{depth} = \lVert \textrm{NR}_d (s, \boldsymbol{R}, \boldsymbol{T}) - \boldsymbol{D} \rVert,\\
\end{equation}

\ourparagraph{Object pose objective.} Combining the mask and depth losses, we obtain the object pose estimation objective:
\begin{equation}
\small
s^*, \boldsymbol{R}^*, \boldsymbol{T}^* = \argmin_{s,\boldsymbol{R}, \boldsymbol{T}}\, L_{mask} + L_{depth},
\label{eq::pose}
\end{equation}
We perform the optimization in the image region centered on the object. We start with a number of randomly initialized poses and select the one that leads to the lowest loss.

\subsection{Joint Optimization}\label{sec_joint_optim}

%So far we described how we optimize hand pose and object pose individually. To obtain the final reconstruction, we optimize their 3D arrangement jointly.
In this section, we describe how to jointly optimize the 3D hand and object results from previous sections (Figure~\ref{fig:pipeline}c). Naively putting them together may result in implausible hand-object reconstructions (Figure~\ref{fig:qual_steps}, row 2), \ie, the hand and object are far away from each other in 3D or having interpenetration. %Obtaining a coherent 3D arrangement of hand and object in one shot is challenging due to the depth and scale ambiguity given only 2D input. 
This issue is caused by the depth and scale ambiguity given only 2D input: a large object distant from the camera can have the same rendering result in 2D as a smaller object closer to the camera. To help resolve the ambiguity, we impose additional constraints based on hand-object distance and collision.

\ourparagraph{Interaction loss.} The reconstructed hand and object could be distant in 3D space. However, when the hand is interacting with objects, their distance should be small. To push the interaction pair closer in 3D, we define an interaction loss based on their chamfer distance.
% COMMENTED OUT
\begin{equation}
\begin{aligned}
\small
    L_{dist} = &\frac{1}{|\boldsymbol{V}_o|} \sum_{x \in \boldsymbol{V}_o} \min_{y \in \boldsymbol{V}_h} \|x - y\|_2 \\
    + & \frac{1}{|\boldsymbol{V}_h|}\sum_{y \in \boldsymbol{V}_h} \min_{x \in \boldsymbol{V}_o} \|x - y\|_2.
\end{aligned}
\end{equation}
For each vertex in the mesh, chamfer distance function finds the nearest point in the other point set, and sums up the distances. We find this loss term to be helpful in correcting the object scale by moving it closer to the hand.

\ourparagraph{Collision loss.} Using the interaction loss alone may result in implausible artifacts, \eg, hand colliding with the object. To resolve the issue, we add an interpenetration loss term to penalize the object vertices that are inside the hand mesh. We use the Signed Distance Field (SDF) from the hand mesh to check if any object vertex is inside the hand. We first calculate a tight box around the hand and voxelize it as a 3D grid for storing the SDF value. %Following~\cite{jiang2020mpshape}, 
We use a modified SDF function $\phi$ for the hand mesh:
\begin{equation}
\small
\phi(\boldsymbol{c}) = -\min(\textrm{SDF}(c_x, c_y, c_z), 0).
\label{eq::sdf1}
\end{equation}
For each voxel cell $\boldsymbol{c} = (c_x, c_y, c_z)$ in the 3D grid, if the cell is inside the hand mesh, $\phi$ takes positive values, proportional to the distance from the hand surface, while $\phi$ is 0 outside of the hand mesh. Then, the interpenetration loss can be calculated as:
\begin{equation}
\small
L_{collision} = \sum_{\boldsymbol{v} \in \boldsymbol{V}_o^*} \phi(\boldsymbol{v}),
\label{eq::sdf2}
\end{equation}
where $\phi(\boldsymbol{v})$ samples the SDF value of each object vertex $\boldsymbol{v}$ from the 3D hand grid in a differentiable way.

\ourparagraph{Joint objective.} By incorporating the loss terms from object pose estimation, we obtain the overall objective for jointly optimizing the hand and the object:
\begin{equation}
\small
L = \lambda_1 L_{mask} + \lambda_2 L_{depth} + \lambda_3 L_{dist} + \lambda_4 L_{collision}.
\end{equation}
%where $\lambda_1$ to $\lambda_4$ are tunable hyperparameters.

%##################################################################################################
\begin{figure}[t]\centering
\includegraphics[width=1.0\linewidth]{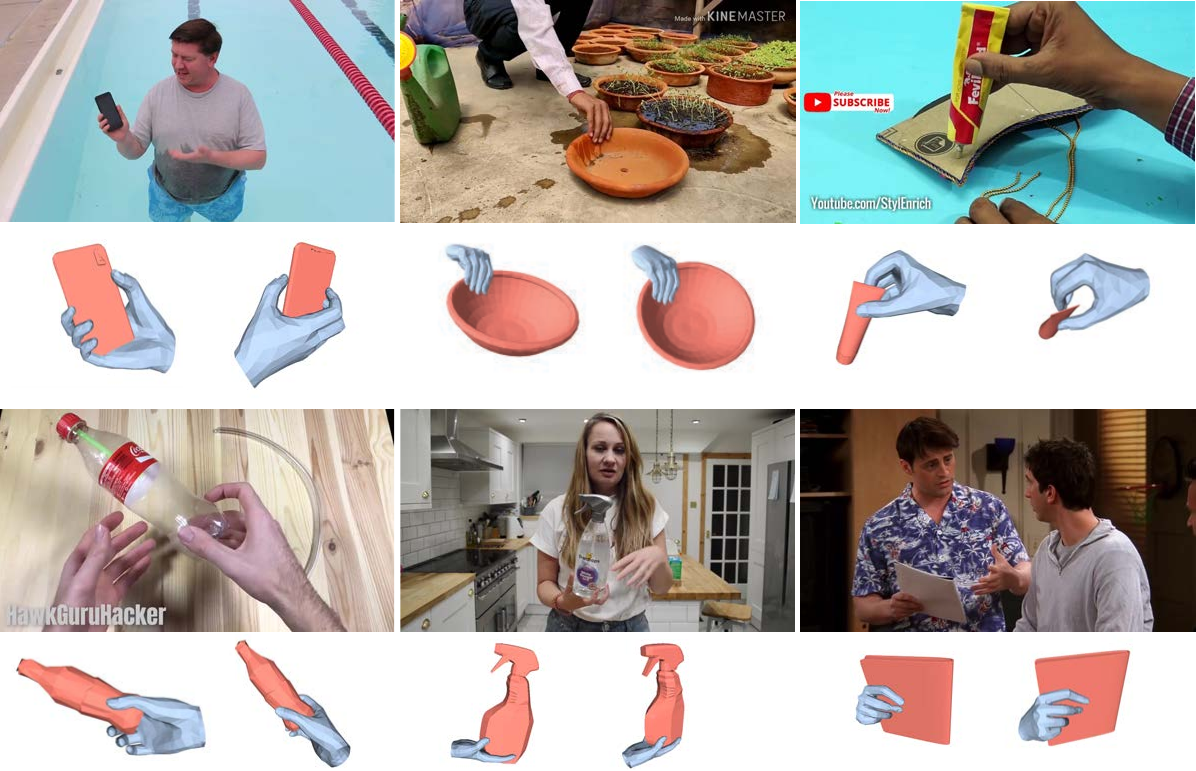}\vspace{0mm}
\caption{\textbf{Qualitative results}. Our method produces reconstructions of reasonably high-quality across a range of viewpoints, activities, and objects.} 
\label{fig:qual_100doh}\vspace{-2mm}
\end{figure}
%##################################################################################################

\subsection{Pose Refinement}

A physically plausible hand-object reconstruction should not only be collision-free, but also have enough hand-object contact area to support the action. However, the interaction loss described in Section~\ref{sec_joint_optim} does not take into account the fine-grained hand-object contact. To further refine the 3D reconstruction, we impose constraints on the hand-object contact as the final step of \method (Figure~\ref{fig:pipeline}d).

Addressing this issue would be easy if we had per-vertex contact area annotation for both hand and object as we could enforce the contact region to be closer. However, obtaining such annotations for large collection of in-the-wild images is challenging. As a more scalable solution, we learn 3D contact priors from a large-scale hand MoCap dataset~\cite{GRAB:2020}. The priors include the region of an object that the person is likely to contact. For example, human is more likely to hold the mug by its handle.

Given the hand mesh and object mesh obtained from the joint optimization, we want to update the hand pose so that it has more reasonable contact with the object. We train a small network to perform hand pose refinement.

The input to the network are the initial hand parameters $(\theta,
\gamma)$ and the distance field $\boldsymbol{F}$ from the hand vertices $\boldsymbol{V}_h^*$ to the object vertices $\boldsymbol{V}_o^*$. For each hand vertex $\boldsymbol{v}_h$, we compute the distance to its nearest object vertex:
\begin{equation}
\small
\boldsymbol{F}(\boldsymbol{v}_h) = \min_{\boldsymbol{v}_o \in \boldsymbol{V}_o^*}\lVert \boldsymbol{v}_h-\boldsymbol{v}_o {\rVert}^2_2
\end{equation}
Then, the network refines the hand parameters $(\boldsymbol{\theta}, \boldsymbol{\beta})$ in an iterative fashion. After each iteration, the distance field between hand and object is updated so that it can be used as input to the next step. The training data is obtained by perturbing the ground-truth hand pose parameters and translation to simulate noisy input estimates.  As shown in Figure~\ref{fig:qual_steps}, we can observe the results after refinement (4th row) can reconstruct more realistic interaction between hand and object than the previous step (3rd row).

%%##################################################################################################
\begin{figure}[t]\centering
\includegraphics[width=1.0\linewidth]{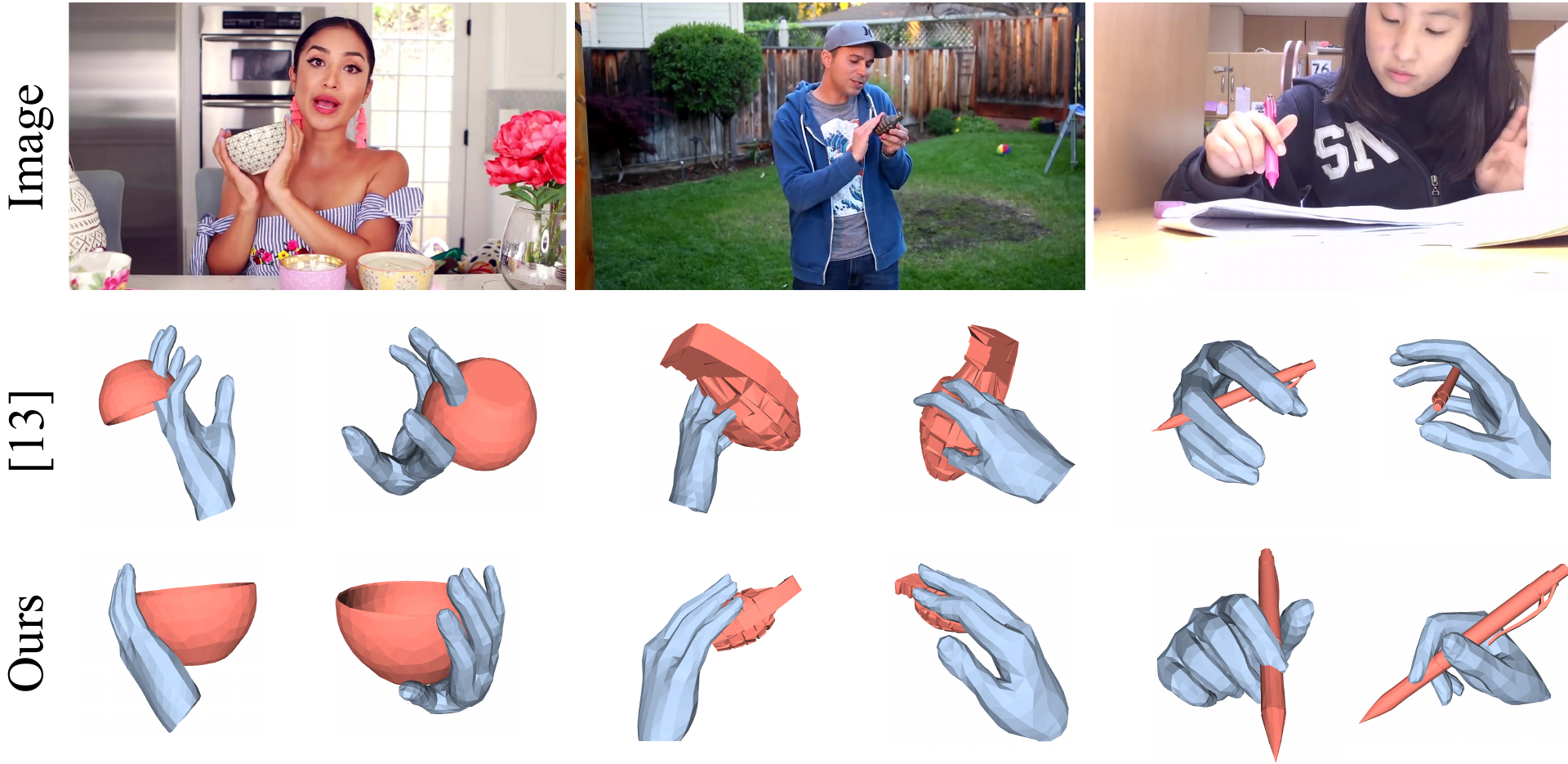}
\caption{\textbf{Qualitative comparison in the wild.} Compared to existing method~\cite{hasson20_handobjectconsist}, our approach produces better hand-object reconstruction across diverse object categories.}
\label{fig:qual_comp}\vspace{-2mm}
\end{figure}
%%##################################################################################################

%%%%%%%%%%%%%%%%%%%%%%%%%%%%%%%%%%%%%%%%%%%%%%%%%%%%%%%%%%%%%%%%%%%%%%%%%%%%%%%%%%%%%%%%%%%%%%%%%%%
\input{Section/experiments}

%##################################################################################################
\begin{figure}[t]\centering
\includegraphics[width=1.0\linewidth]{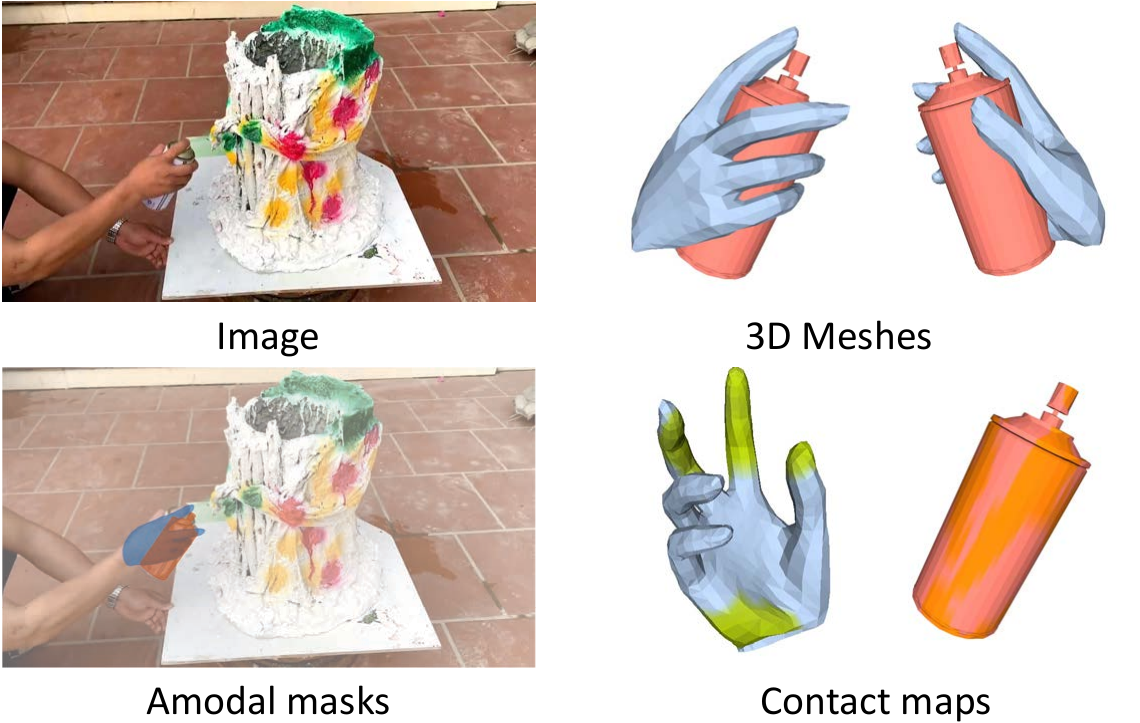}\vspace{-0mm}
\caption{\textbf{Example annotations.} We use the techniques proposed in this work to annotate in-the-wild images and obtain 3D meshes, amodal masks, and contact maps.}
\label{fig:labels}\vspace{-3mm}
\end{figure}
%##################################################################################################

%%%%%%%%%%%%%%%%%%%%%%%%%%%%%%%%%%%%%%%%%%%%%%%%%%%%%%%%%%%%%%%%%%%%%%%%%%%%%%%%%%%%%%%%%%%%%%%%%%%
\section{Dataset}\label{sec:data}

We describe our dataset collection procedure and present analysis that highlights the variety our data.

\ourparagraph{Image collections.} As a source of in the wild data we use static frames from the EPIC Kitchens~\cite{Damen2018, Damen2020} and the 100 Days of Hands~\cite{Shan2020} datasets, noting that we do not exploit any temporal information. These datasets contain a range of interesting hand-object interaction scenarios with varied objects, people, and viewpoints (both first- and third-person). To determine candidate images for reconstruction, we use a hand and object detector~\cite{Shan2020} and select images that contain a high bounding box overlap between an object and a hand.

\subsection{Dataset Construction}

Our annotation procedure consists of three steps: selecting a 3D object model, performing reconstruction using the method proposed in \S\ref{sec:method}, and verification of the results.

\ourparagraph{Step 1: Model selection.} The first step of our annotations requires the annotator to choose an appropriate 3D object model for the object being manipulated by the hand. We maintain a collection of available object models. If the required object is already present in the collection, the annotator directly selects the model. If not, the annotator finds an appropriate model online and adds it to the collection. Two primary sources of 3D object models we use are the YCB dataset~\cite{calli2015ycb} and the Free3D online platform.

%##################################################################################################
\begin{table}[t]\centering
\resizebox{\columnwidth}{!}{\tablestyle{8pt}{1.05}
\begin{tabular}{@{}l|cc|cc@{}}
    & Object IoU  & Hand IoU &  Quality & Match? \\\shline
 Large & 0.84 & 0.67 & - & -\\
 Medium & 0.78 & 0.69 & - & -\\
 Small & 0.64 & 0.63 & - & - \\
 All & 0.77 & 0.68 & 4.16 & 92\%  \\
\end{tabular}}
\vspace{-2mm}
\caption{\textbf{Dataset evaluation.} \emph{Left:} Amodal masks derived from our 3D annotations have a high overlap with ground truth amodal masks labeled by humans. \emph{Right:} Users, asked to rate the quality of our 3D annotations from 1 to 5, find that they are of good quality on average and include a 3D object model that matches the true object in most cases.}
\label{tab:dataset:eval}\vspace{-2mm}
\end{table}
%##################################################################################################

%%##################################################################################################
\begin{figure*}[t]\centering
\includegraphics[width=1.0\linewidth]{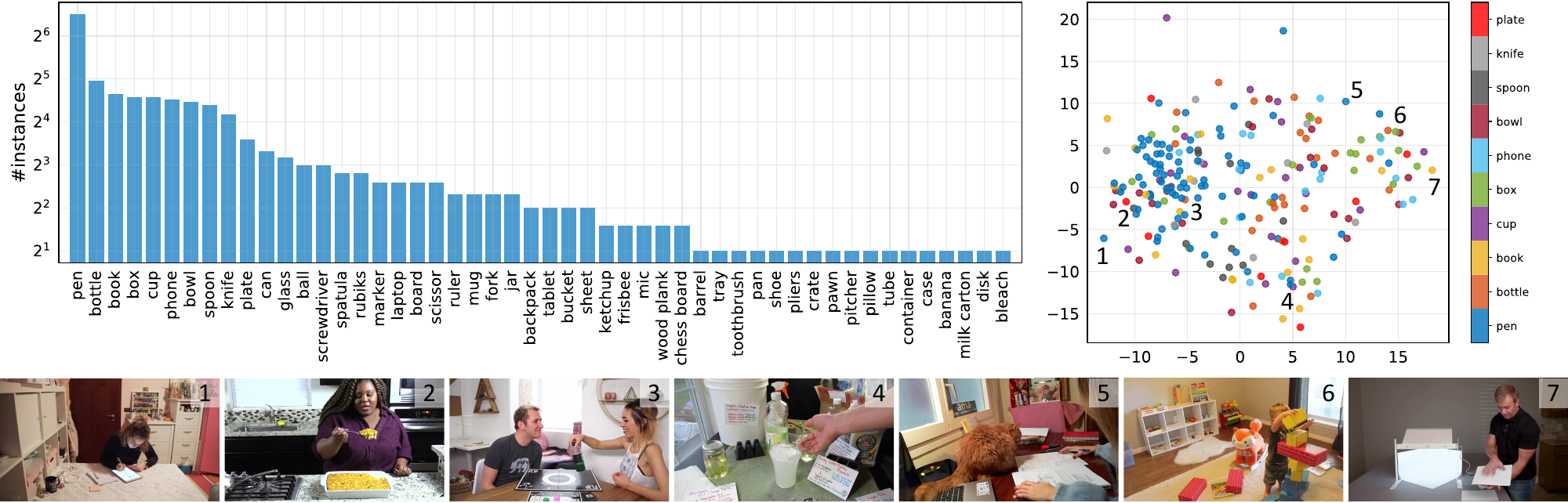}
\caption{\textbf{Variety of objects and grasps.} We present analysis that shows the variety of objects and grasp types in our data. \emph{Top left:} Our data contains 121 object categories and a total of 500 instances. The object distribution has a long tail. \emph{Top right:} We embed 3D hand poses into 2D space using Isomap~\cite{Tenenbaum2000}. Each point is a hand-object interaction and is color-coded by object category. We notice that there is a cluster of pens but no other clear clusters. This suggests that our data contains a variety of grasp types for each object category rather than only iconic grasps. Indeed, we see examples of different object categories with similar grasp types (pen and spoon) and same object category with different grasp types (pen). \emph{Bottom:} We observe that the first embedding dimension (x axis) corresponds to the closure of the grasp. We show examples for increasing value of x. From left to right, we see that the grasps gradually transitions from fully closed to fully open.}
\label{fig:dist_emb}\vspace{-2mm}
\end{figure*}
%%##################################################################################################

\ourparagraph{Step 2: Reconstruction.} Next, we perform the hand-object reconstruction using our method, called \method, proposed in \S\ref{sec:method}. This step is semi-automatic and relies on the annotator to select the appropriate loss weights to obtain a good reconstruction. In practice, most annotators find that our default loss settings lead to a reasonable starting point.

\ourparagraph{Step 3: Verification.} In practice, we find that \method results in good reconstructions in many cases. However, there are still cases where the results are imperfect across different viewpoints due to ambiguity. Thus, to ensure good annotation quality, we perform an additional step and verify the reconstructions obtained in step 2. Specifically, we ask the annotator to inspect the result from step 2 and take one of three possible actions: accept it if good, return it to step 2 if promising, and remove it from consideration if unlikely to improve. We iterate back and forth with step 2 until we converge to a set of reconstructions of reasonable quality.

\ourparagraph{Summary.} To summarize, the output of our annotation procedure is that for each image we have: 3D object model, 3D object pose, and 3D hand pose. Moreover, we can easily derive additional annotations, such as amodal masks or contact maps. Example annotations are shown in Figure~\ref{fig:labels}.

\subsection{Dataset Evaluation}

Annotating data in 3D is hard. Evaluating the quality of annotations is harder. To judge the quality of the collected annotations, we use two types of evaluation. The evaluation is performed on a sample set of 100 images.

\ourparagraph{Amodal mask accuracy.} To evaluate our annotations, we require a signal that is predictive of reconstruction quality and can be labeled reliably by humans. Amodal instance masks, that include both visible and occluded parts of the object~\cite{Li2016}, are a good fit. Given only the visible portions of the image, there are many plausible configurations for the hidden object parts, especially for articulated objects like hands. Nevertheless, humans are capable of predicting occluded regions with high consistency~\cite{Zhu2017}.

We ask human annotators to label amodal masks for hands and objects, which serves as ground truth. We then compare amodal masks derived from our reconstructions to the ground truth. In Table~\ref{tab:dataset:eval}, we report the mean intersection (IoU) scores for the hands and the objects. Similar to~\cite{Lin2014}, we show results for different object sizes. We observe that our amodal masks have a high overlap with the ground truth. As expected, the overlaps are higher for larger objects.

\ourparagraph{User study.} We also perform a user study. Given the input image and the annotated hand-object reconstruction, we ask the users to assign a quality score to each example on a scale of 1 to 5. The users are instructed to assign 1 when the reconstruction is poor (\eg heavy collision or hand being far from the object) and 5 when there are no clear imperfections visible. The users can rotate the result in 3D visualization to view from different angles. We also ask the users to say if the object in the image matches the 3D model.

In Table~\ref{tab:dataset:eval}, we report the results. The average reconstruction quality we obtain is 4.16. This suggests that most of our annotations are of good quality. Moreover, we find that the 3D object model matches the true object in 92\% of the cases. Among the 8\%, most are due to imprecise mesh topology, \eg a cylinder fitted to a mug with a handle.

\subsection{Dataset Analysis}\label{sec:dataset:analysis}

We annotated 500 images using the proposed procedure. We now present the analysis of the collected data.

\ourparagraph{Object variety.} Our dataset contains 121 object categories covering a wide variety of daily objects. In Figure~\ref{fig:dist_emb}, top-left, we show the object distribution for the 50 most frequent objects. There are some categories with many examples and a long tail of object categories with few examples.

\ourparagraph{Grasp variety.} A unique feature of our data is that it provides a variety of hand-object interactions \emph{in the wild}. This allow us to study and learn about human grasps using real-world data. In Figure~\ref{fig:dist_emb}, top-right, we embed 3D hand poses into 2D space using Isomap~\cite{Tenenbaum2000}. Each point corresponds to an interaction and is color-coded by object category.

We observe that there is a cluster of pens on the left but no other clear clusters. This suggests that our data contains a variety of grasp types for each object category, rather than only iconic grasps. Indeed, looking closer we notice that there are many examples of similar grasps for different object categories (\eg pen and spoon) as well as different grasp types for the same object category (\eg pen).

\ourparagraph{Grasp structure.} We further observe an interesting pattern in the data. In particular, we find that the first dimension of the hand pose embedding (x axis) corresponds to the closure of the hand. In Figure~\ref{fig:dist_emb}, bottom, we show example images for increasing value of x. We see that the grasps gradually transition from fully closed to fully open.

%%%%%%%%%%%%%%%%%%%%%%%%%%%%%%%%%%%%%%%%%%%%%%%%%%%%%%%%%%%%%%%%%%%%%%%%%%%%%%%%%%%%%%%%%%%%%%%%%%%
\section{Conclusion}

In this work, we propose a new technique for reconstructing hand-object interactions with the help of 2D image cues and 3D contact priors. We use this technique in conjunction with human intervention to construct a new 3D hand-object interaction dataset in the wild. We encourage the readers to see the \href{https://people.eecs.berkeley.edu/~zhecao/rhoi}{project page} for additional materials.

\ourparagraph{Acknowledgments.} This work was supported by DARPA program on Machine Common Sense.

%%%%%%%%%%%%%%%%%%%%%%%%%%%%%%%%%%%%%%%%%%%%%%%%%%%%%%%%%%%%%%%%%%%%%%%%%%%%%%%%%%%%%%%%%%%%%%%%%%%
\section*{Appendix}

In this appendix we provide more information about 2D object masks, the depth loss, and implementation details. We further supplement the results from the main text with failure cases and additional qualitative examples.

%%##################################################################################################
\begin{figure}[t]\centering
\includegraphics[width=0.49\linewidth]{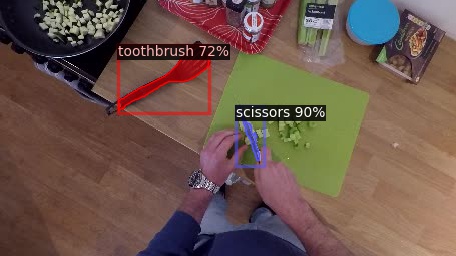}\hspace{0.5mm}
\includegraphics[width=0.49\linewidth]{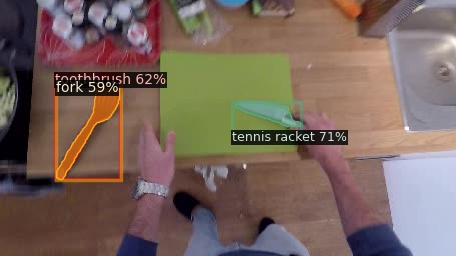}
\caption{\textbf{Unlabeled object categories.} We are interested in reconstructing object categories that are not labeled in existing image segmentation datasets. Our key insight that allows us to do this is that the models trained on large labeled datasets can produce reasonable object masks even if the category predictions are incorrect, as shown here.}
\label{fig:mask_cat}%\vspace{-2mm}
\end{figure}
%%##################################################################################################

\subsection*{Object Masks}

In Figure~\ref{fig:mask_cat}, we show example model predictions for object categories that are not labeled in existing object segmentation datasets. We see that even if the predicted categories are incorrect, the object masks are quite reasonable. For example, the model does not know what a spatula is called but is still able to segment it. We use this observation in our method to reconstruct a range of object categories.

%##################################################################################################
\begin{figure}[t]\centering
\includegraphics[width=1.0\linewidth]{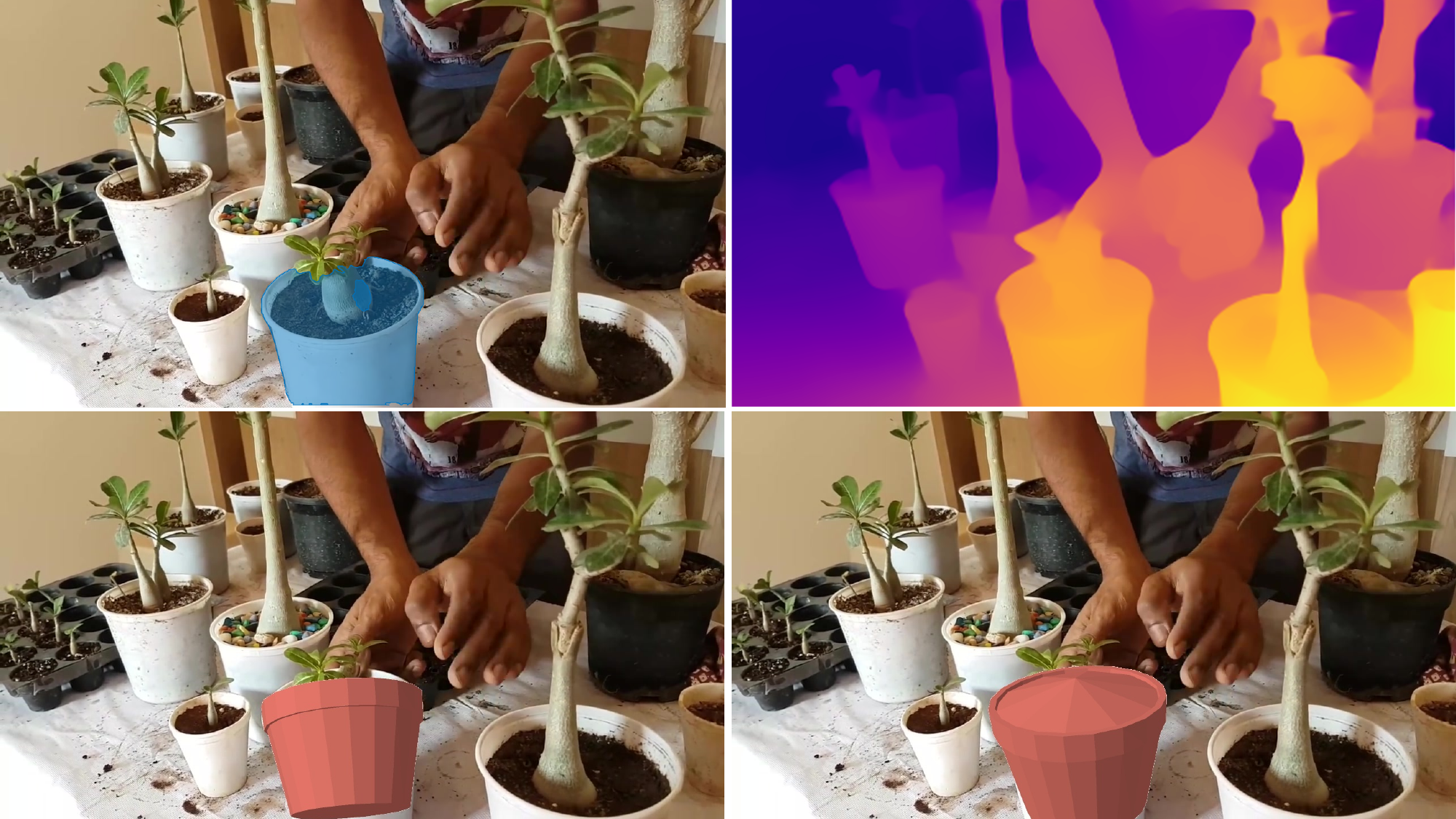}\vspace{-1mm}
\caption{\textbf{The effect of the depth loss.} We show the predicted 2D object mask visualized in the image (top-left), the predicted depth map (top-right), the object pose estimation result without depth loss (bottom-left), and the result with depth loss (bottom-right).}
\label{fig:qual_depth}\vspace{-2mm}
\end{figure}
%##################################################################################################

\subsection*{Depth Loss}

In Figure~\ref{fig:qual_depth}, we demonstrate the effect of the depth loss for object pose estimation. 2D mask loss is insufficient in some cases, it does not capture geometry information and can be ambiguous---multiple object poses can lead to similar 2D masks as shown in the two results (the bottom row). To resolve the ambiguity, we introduce a new depth loss which compares the difference between the rendered object depth map and the estimated depth map from off-the-shelf method. As shown in our final result, this additional loss enables our method to achieve more accurate estimation of the object orientation.

\subsection*{Implementation Details}\label{sec_imp_details}

To speed up optimization, we preprocess our 3D object models to have $\app$800 faces. For object pose estimation, we use $1000$ initializations for object rotation sampled uniformly from $[-\pi, \pi]$. Object translation is zero initialized. We select the object pose with the lowest loss. When optimizing the hand pose, we weight the 2D keypoint loss $L_{2D}$ using the confidence score of each predicted hand joint. For the joint optimization of hand and object, our default loss weights are as follows: $\lambda_1 = 200, \lambda_2 = 20, \lambda_3 = 1000, \lambda_4 = 5$.  When computing the interpenetration loss between hand and object, we voxelize the hand mesh into a $32 \times 32 \times 32$ grid. For the refinement networks, we use $10,000$ object vertices to compute the distance field $\boldsymbol{D}$ from hand to object and perform 3 refinement iterations. In practice, it takes $\app$1 minute to run our method on an image using a GTX 2080 Ti GPU, including $\app$18 seconds for optimizing the object pose and $\app$30 seconds for the joint optimization.

%##################################################################################################
\begin{figure}[t]\centering
\includegraphics[width=1.0\linewidth]{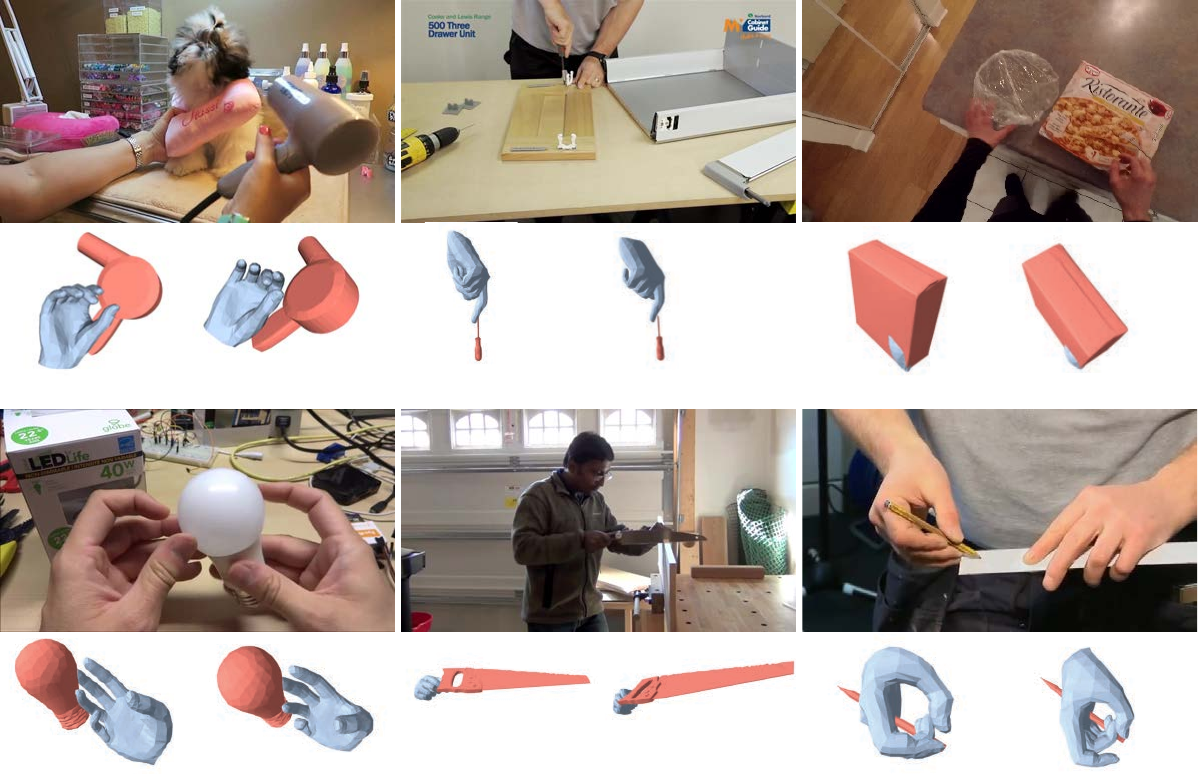}\vspace{-1mm}
\caption{\textbf{Failure cases.} We show failure cases of different parts of our method: hand pose estimation (left), object pose estimation (middle), and the joint optimization (right).}
\label{fig:qual_failure}\vspace{-2mm}
\end{figure}
%##################################################################################################

\subsection*{Failure Cases}
 
In Figure~\ref{fig:qual_failure}, we show the representative failure modes of our method. In the left column, we observe that the hand pose estimation fails when the hand is highly occluded or cropped by the image. In the middle column, we see incorrect object pose predictions due to the ambiguity of the 2D mask (top) and the imperfect object model (bottom). In the right column, we observe examples where the joint optimization converges to an undesirable local minima.

%##################################################################################################
\begin{figure*}[t]\centering
\includegraphics[width=1.0\linewidth]{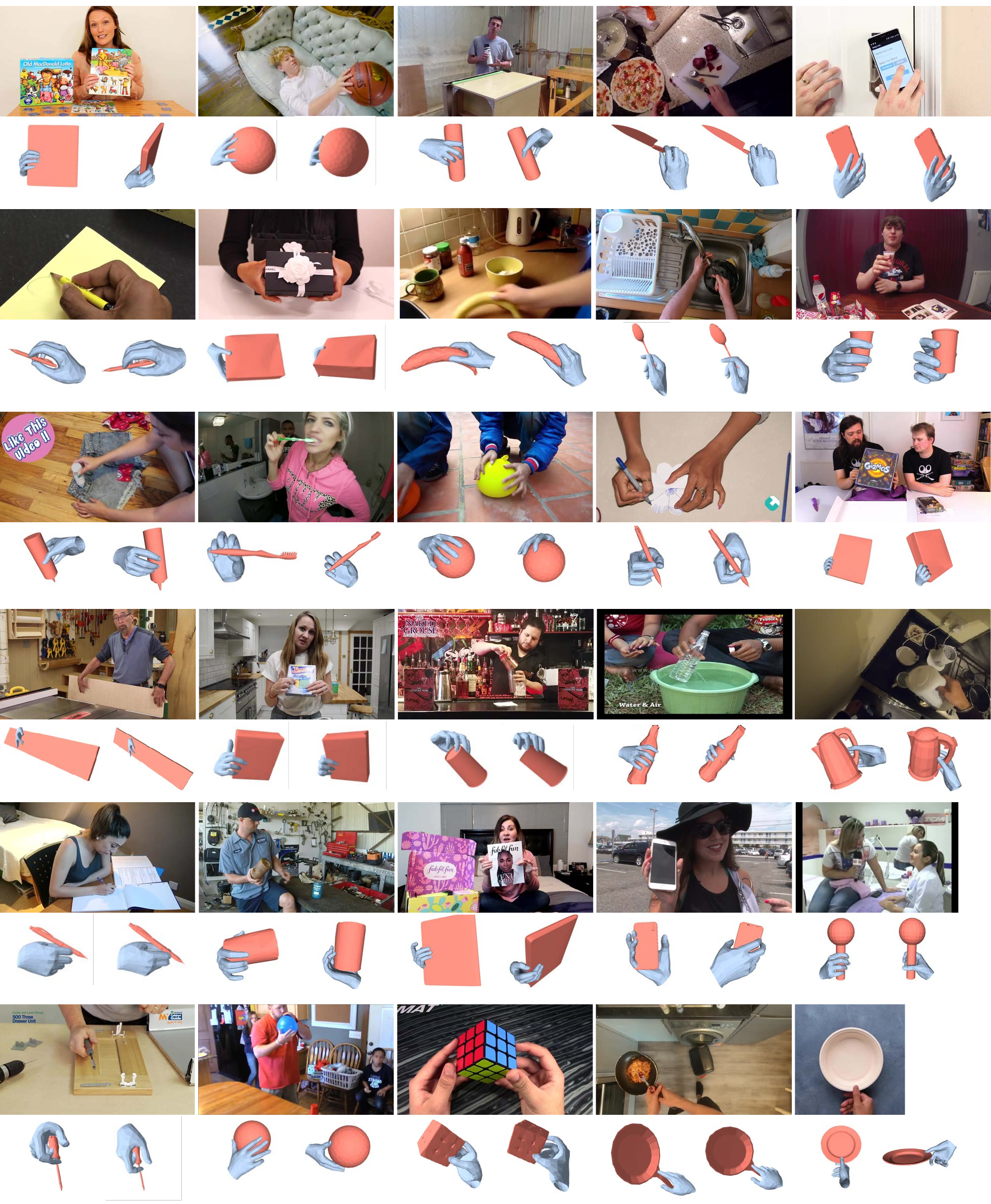}\vspace{-0mm}
\caption{\textbf{Additional qualitative results.} Our method, \method, produces strong results for a range of interactions and objects.}
\label{fig:qual_additional}\vspace{-2mm}
\end{figure*}
%##################################################################################################

%%%%%%%%%%%%%%%%%%%%%%%%%%%%%%%%%%%%%%%%%%%%%%%%%%%%%%%%%%%%%%%%%%%%%%%%%%%%%%%%%%%%%%%%%%%%%%%%%%%
\clearpage
{\small\bibliographystyle{ieee_fullname}\bibliography{references}}

\end{document}

%% file: Section/related_work.tex
\section{Related Work}

%##################################################################################################
\begin{figure*}[!ht]\centering
\includegraphics[width=1.0\linewidth]{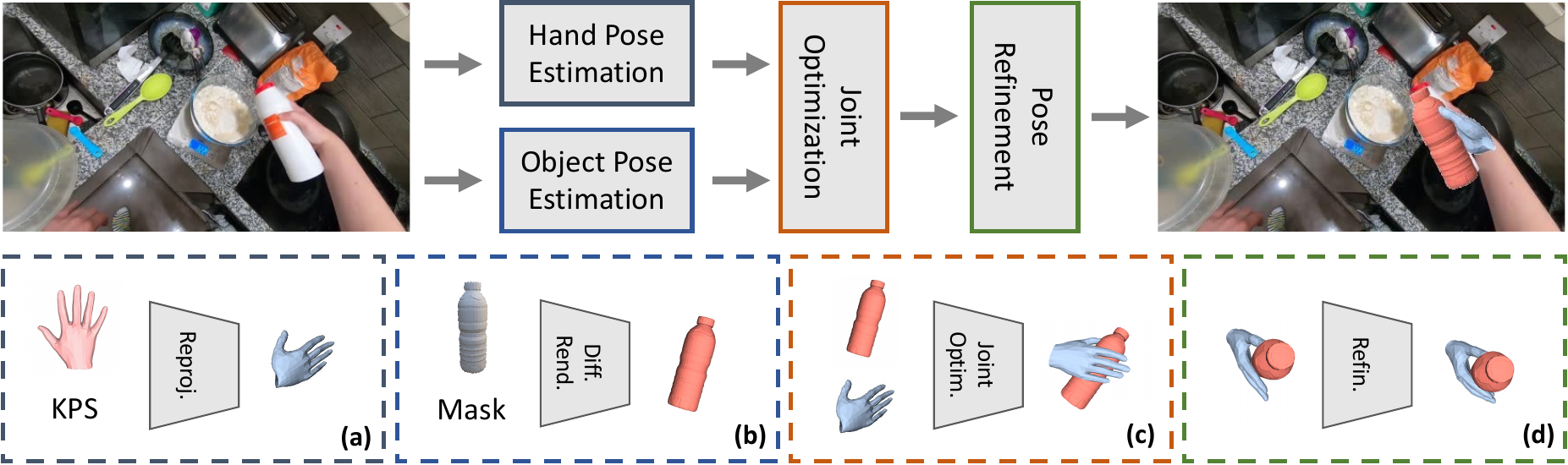}\vspace{-0mm}
\caption{\textbf{Method.} In this work, we propose an optimization-based method, called \emph{\method}, that leverages 2D image cues and 3D contact priors for reconstructing hand-object interactions in the wild. It consists of four steps: (a) hand pose estimation by 2D keypoints fitting, (b) object pose estimation via differentiable rendering, (c) joint optimization for hand-object configuration, and (d) hand pose refinement using 3D contact priors learnt from MoCap data.}
\label{fig:pipeline}\vspace{-2mm}
\end{figure*}
%##################################################################################################

\ourparagraph{3D hand pose estimation.} Many previous works on hand pose estimation directly predict 3D joint locations from either depth~\cite{sharp2015accurate, sridhar2013interactive, tagliasacchi2015robust, tzionas2016capturing, ye2016spatial, yuan2018depth, moon2018v2v} or RGB~\cite{romero2010hands, mueller2018ganerated, cai2018weakly, yang2019disentangling, zimmermann2017learning} images. Some recent works predict 3D hand joint rotations and shape parameters of parametric hand models such as MANO~\cite{romero2017embodied}. Fitting-based approaches~\cite{SMPL-X:2019, xiang2019monocular, Kulon_2020_CVPR} fit such parametric models to 2D keypoint detections to optimize 3D hand parameters. Learning-based approaches~\cite{zhou2020monocular, rong2020frankmocap} utilize deep networks to directly predict the hand parameter from RGB image input. Recently, \cite{Kulon2019, Kulon_2020_CVPR} proposes to use mesh convolution to directly predict 3D hand mesh reconstruction. We use a learning-based method~\cite{rong2020frankmocap} to obtain the initial hand pose estimation and further improve the result by imposing constraints on 2D hand keypoints and 3D hand-object contact priors.

\ourparagraph{3D object pose estimation.} There are many existing works on estimating 3D object pose from a single image. Some approaches~\cite{factored3dTulsiani17, kundu20183d, Gkioxari2019, kuo2020mask2cad} utilize neural network to predict the object shape, translation, and global orientation in the camera coordinate. These methods are trained with limited object categories and have difficulty generalizing to new objects. On the other hand, some approaches~\cite{lim2013parsing, Michel2017, Xiang2018, zhang2020phosa, Sun2018, Sahasrabudhe2019} assume known 3D object model and focus on 6 DOF object pose prediction. In this work, we take a fitting-based approach similar to~\cite{Sun2018, zhang2020phosa}. Our main novelty is the usage of a depth loss term which improves the results by imposing object shape constraints.

\ourparagraph{3D hand and object pose estimation.} Early approaches in modeling hand and object require the input of multi-view image~\cite{oikonomidis2011full} or RGB-D sequence~\cite{tzionas2016capturing}. Recently, \cite{hasson19_obman} proposes a deep model trained on synthetic data to reconstruct hand and object meshes from a monocular RGB image. ~\cite{tekin2019h} designs a neural network to jointly predict 3D hand pose and 3D object bounding boxes with a focus on egocentric scenarios. \cite{hasson20_handobjectconsist} proposes to leverage photometric consistency from temporal frames as additional signal for training the model with sparse set of annotated data. All these approaches were trained and tested on in-the-lab or synthetic datasets. In this work, we propose an approach without 3D supervision and we are the first to achieve good hand-object results in the wild from a single image.

\ourparagraph{3D hand-object datasets.} Datasets of hand grasping scenario usually require manual annotations~\cite{sridhar2016real, corona2020ganhand} or depth tracking ~\cite{tzionas2016capturing} to obtain the ground truth. To avoid the manual efforts, ~\cite{hernando2018cvpr} uses motion capture system with magnetic sensors to collect annotations. \cite{hasson19_obman} uses simulation to collect a synthetic hand-object dataset. \cite{zimmermann2019freihand, hampali2020honnotate} introduces large-scale dataset with 3D annotation optimized from multi-view setups. Some recent datasets~\cite{Brahmbhatt_2019_CVPR, Brahmbhatt_2020_ECCV,GRAB:2020} also provide annotation for hand-object contact area in addition to 3D hand pose and object pose. The contact area is collected from either thermal sensor~\cite{Brahmbhatt_2019_CVPR, Brahmbhatt_2020_ECCV} or marker-based MoCap system.  All these datasets are of great efforts in modeling 3D hand-object interaction, however, they can only be collected in the lab setting due to the specific camera setup. As a result, limited number of participants and objects are present in them (as in Table~\ref{tab:datasets}). In this work, we contribute a dataset with in-the-wild images and diverse object categories. 3D annotations are obtained by running our optimization-based method \emph{and} human intervention to achieve high quality.

\ourparagraph{Optimizing 3D interactions.} Our method is in line with recent optimization-based approaches for modeling 3D interactions between human and scene~\cite{PROX:2019}, human and objects~\cite{zhang2020phosa}, and among multiple persons~\cite{jiang2020mpshape}. To obtain good 3D reconstructions, these methods require extra 3D input. For example, \cite{PROX:2019} requires the input of full 3D reconstructed scene mesh to impose geometry constraints. \cite{zhang2020phosa} requires manually labeling human-object mesh vertices for fine-grained interaction pairs and is only applied to 8 object categories in COCO~\cite{Lin2014}. In this work, we focus on modeling hand-object interactions. Our key advantage is the ability to deal with diverse objects in the wild without extra input. We propose to model contact priors using a scalable data-driven approach that levarages the available 3D MoCap data. Together with a new method to obtain object masks, our approach is shown to be able to reconstruct hand interactions with 121 different object categories.

%% file: Section/experiments.tex
\section{Method Evaluation}

In this section, we compare our method to existing methings in two settings: quantitatively in the lab and qualitatively in the wild. We further present ablation studies of different aspect of our method.

\subsection{Quantitative Comparison in the Lab}
In Table~\ref{tab:comparison:lab}, we perform quantitative evaluation of our method in the HO3D dataset~\cite{hampali2020honnotate} and FPH dataset~\cite{hernando2018cvpr}. HO3D~\cite{hampali2020honnotate} dataset contains 3D annotations for both the hand and object of 68 video sequences, 10 subjects, and 10 objects. FPHA~\cite{hernando2018cvpr} dataset utilizes a MoCap system to capture hand-object interaction. 3D object pose annotations are available for 4 objects and subset of videos. We follow the same testing split as~\cite{hasson20_handobjectconsist} for comparison.

\ourparagraph{Method for comparison} 
We compare against the state-of-the-art (SOTA) approach~\cite{hasson20_handobjectconsist} with the same input of monocular RGB image and the known 3D object model. \cite{hasson20_handobjectconsist} uses a feed-forward neural network to predict 3D hand pose and object pose where its single-frame model with full 3D supervision shows SOTA performance.

\ourparagraph{Evaluation metric.}  We report the mean average error (MAE) over 21 hand joints. The error measures the Euclidean distance between predictions and ground truth. Following~\cite{hampali2020honnotate}, we calculate the error after aligning hand root position and global scale with the ground-truth.

For evaluating object pose, we calculate the Chamfer distance between ground-truth object vertices and predicted object vertices (obtained by rotating the input CAD model with the predicted object pose).

\ourparagraph{Results.} 
Table~\ref{tab:comparison:lab} shows our method achieves better accuracy than~\cite{hasson20_handobjectconsist} in 3D hand and object error. In HO3D dataset (left table), our predictions have smaller hand joint error of $9.7$ mm \vs $14.7$ mm and smaller object Chamfer distance of $19.9$ \vs $26.8$.
In FPHA dataset (right table), our method achieves smaller hand joints error ($14.2$ mm \vs $18.0$ mm). Our object error is slightly larger than \cite{hasson20_handobjectconsist}. The main reason is that \cite{hasson20_handobjectconsist} uses the action split of FPHA, i.e., same objects with different action labels are used for training and testing. In comparison, our method are tested directly without 3D supervision in those datasets. 

%##################################################################################################
\begin{table}[t]\centering
\resizebox{.46\columnwidth}{!}{\tablestyle{8pt}{1.05}
\begin{tabular}{@{}l|cc@{}}
 Metrics & \cite{hasson20_handobjectconsist} & Ours \\\shline
 Hand MAE $^{\downarrow}$  & 14.7  & \textbf{9.7}  \\
 Obj CF dist $^{\downarrow}$    & 26.8  & \textbf{19.9}  
 %$F_1$@30mm $^{\uparrow}$    & 62.0  &\textbf{82.9}    \\
\end{tabular}}\hspace{5mm}
\resizebox{.46\columnwidth}{!}{\tablestyle{8pt}{1.05}
\begin{tabular}{@{}l|cc@{}}
 Metrics & \cite{hasson20_handobjectconsist} & Ours \\\shline
 Hand MAE $^{\downarrow}$  & 18.0  &  \textbf{14.2} \\
 Obj MAE $^{\downarrow}$    & \textbf{22.3}  &  23.9 
\end{tabular}}
\vspace{-1mm}
\caption{\textbf{Quantitative comparison in the lab}. Our method achieves results better or on par with the state of the art on popular in-the-lab datasets: HO3D (left) and FPHA (right).}
\label{tab:comparison:lab}%\vspace{-2mm}
\end{table}
%##################################################################################################

%##################################################################################################
\begin{table}[t]\centering
\resizebox{\columnwidth}{!}{\tablestyle{10pt}{1.05}
\begin{tabular}{@{}l|cc@{}}
    & HO Distance $^{\downarrow}$  & Collision  Score $^{\downarrow}$ \\\shline
 Individual results &  414.8 & 0 \\
 + Interaction loss  & 71.5 & 39.8 \\
 + Depth loss & 75.2 & 17.6 \\
 + Penetration loss & 76.4 & 7.7 \\
 + Refinement &  75.8 & 6.5 \\
\end{tabular}}\vspace{-1mm}
\caption{\textbf{Ablations on loss terms and pose refinement.} From top to bottom, we add each component one by one (cumulative) and evaluate the prediction in terms of the distance between hand and object, and the collision score.}
\label{tab:ablation:loss}\vspace{-2mm}
\end{table}
%##################################################################################################

\subsection{Qualitative Comparison in the Wild}

In Figure~\ref{fig:qual_comp}, we show side-by-side qualitative comparisons with~\cite{hasson20_handobjectconsist} using in-the-wild images from~\cite{Shan2020}, which clearly shows the advantage of our method. Though~\cite{hasson20_handobjectconsist} achieves good performance in the lab, it struggles on in-the-wild images. This was primarily due to the lack of labeled in-the-wild training data with diverse object categories. As a result, the model trained on limited object categories in the lab has difficulty in generalizing to new unseen objects.

\subsection{Ablation Studies}

We now present the ablation studies of the method. We evaluate the influence of the joint optimization loss terms and the refinement stage on the overall results. We report the distance between the estimated hand and object centers and the collision score computed based on SDF. The more the object intersects with the hand the larger the collision score is. A good reconstruction should have small collision and small hand-object distance.

In Table~\ref{tab:ablation:loss}, we observe that the individually reconstructed hand and object are far from each other, resulting in large distance and no collision. By adding the interaction loss, the distance decreases quickly to $71.5$ mm but also results in a large collision, i.e, $39.8$. Adding the depth and collision losses reduces the collision score to $7.7$ while keeping a similar hand-object distance, i.e., $76.4$ mm. The refinement stage makes small adjustment to the final result and can slightly reduce both the collision score ($6.5$ \vs $7.7$) and hand-object distance ($75.8$ mm \vs $76.4$ mm). These findings are consistent with visualization in Figure~\ref{fig:qual_steps}.

%% file: main.bbl
\begin{thebibliography}{10}\itemsep=-1pt

\bibitem{Brahmbhatt_2019_CVPR}
Samarth Brahmbhatt, Cusuh Ham, Charles~C. Kemp, and James Hays.
\newblock {ContactDB}: Analyzing and predicting grasp contact via thermal
  imaging.
\newblock In {\em CVPR}, 2019.

\bibitem{Brahmbhatt_2020_ECCV}
Samarth Brahmbhatt, Chengcheng Tang, Christopher~D. Twigg, Charles~C. Kemp, and
  James Hays.
\newblock {ContactPose}: A dataset of grasps with object contact and hand pose.
\newblock In {\em ECCV}, 2020.

\bibitem{cai2018weakly}
Yujun Cai, Liuhao Ge, Jianfei Cai, and Junsong Yuan.
\newblock Weakly-supervised 3d hand pose estimation from monocular rgb images.
\newblock In {\em ECCV}, 2018.

\bibitem{calli2015ycb}
Berk Calli, Arjun Singh, Aaron Walsman, Siddhartha Srinivasa, Pieter Abbeel,
  and Aaron~M Dollar.
\newblock The ycb object and model set: Towards common benchmarks for
  manipulation research.
\newblock In {\em ICRA}, 2015.

\bibitem{8765346}
Zhe {Cao}, Gines {Hidalgo Martinez}, Tomas {Simon}, Shih-En. {Wei}, and Yaser
  {Sheikh}.
\newblock Openpose: Realtime multi-person 2d pose estimation using part
  affinity fields.
\newblock {\em TPAMI}, 2019.

\bibitem{corona2020ganhand}
Enric Corona, Albert Pumarola, Guillem Alenya, Francesc Moreno-Noguer, and
  Gr{\'e}gory Rogez.
\newblock Ganhand: Predicting human grasp affordances in multi-object scenes.
\newblock In {\em CVPR}, 2020.

\bibitem{Damen2020}
Dima Damen, Hazel Doughty, Giovanni~Maria Farinella, Antonino Furnari,
  Evangelos Kazakos, Jian Ma, Davide Moltisanti, Jonathan Munro, Toby Perrett,
  Will Price, et~al.
\newblock Rescaling egocentric vision.
\newblock {\em arXiv:2006.13256}, 2020.

\bibitem{Damen2018}
Dima Damen, Hazel Doughty, Giovanni Maria~Farinella, Sanja Fidler, Antonino
  Furnari, Evangelos Kazakos, Davide Moltisanti, Jonathan Munro, Toby Perrett,
  Will Price, et~al.
\newblock Scaling egocentric vision: The epic-kitchens dataset.
\newblock In {\em ECCV}, 2018.

\bibitem{hernando2018cvpr}
Guillermo Garcia-Hernando, Shanxin Yuan, Seungryul Baek, and Tae-Kyun Kim.
\newblock First-person hand action benchmark with {RGB-D} videos and {3D} hand
  pose annotations.
\newblock In {\em CVPR}, 2018.

\bibitem{Gibson1977}
James~J Gibson.
\newblock The theory of affordances.
\newblock {\em Hilldale, USA}, 1977.

\bibitem{Gkioxari2019}
Georgia Gkioxari, Jitendra Malik, and Justin Johnson.
\newblock Mesh r-cnn.
\newblock In {\em ICCV}, 2019.

\bibitem{hampali2020honnotate}
Shreyas Hampali, Mahdi Rad~Markus Oberweger, and Vincent Lepetit.
\newblock Honnotate: A method for 3d annotation of hand and object poses.
\newblock In {\em CVPR}, 2020.

\bibitem{PROX:2019}
Mohamed Hassan, Vasileios Choutas, Dimitrios Tzionas, and Michael~J. Black.
\newblock Resolving {3D} human pose ambiguities with {3D} scene constraints.
\newblock In {\em ICCV}, 2019.

\bibitem{hasson20_handobjectconsist}
Yana Hasson, Bugra Tekin, Federica Bogo, Ivan Laptev, Marc Pollefeys, and
  Cordelia Schmid.
\newblock Leveraging photometric consistency over time for sparsely supervised
  hand-object reconstruction.
\newblock In {\em CVPR}, 2020.

\bibitem{hasson19_obman}
Yana Hasson, G{\"u}l Varol, Dimitris Tzionas, Igor Kalevatykh, Michael~J.
  Black, Ivan Laptev, and Cordelia Schmid.
\newblock Learning joint reconstruction of hands and manipulated objects.
\newblock In {\em CVPR}, 2019.

\bibitem{jiang2020mpshape}
Wen Jiang, Nikos Kolotouros, Georgios Pavlakos, Xiaowei Zhou, and Kostas
  Daniilidis.
\newblock Coherent reconstruction of multiple humans from a single image.
\newblock In {\em CVPR}, 2020.

\bibitem{kato2018renderer}
Hiroharu Kato, Yoshitaka Ushiku, and Tatsuya Harada.
\newblock Neural 3d mesh renderer.
\newblock In {\em CVPR}, 2018.

\bibitem{kirillov2020pointrend}
Alexander Kirillov, Yuxin Wu, Kaiming He, and Ross Girshick.
\newblock Pointrend: Image segmentation as rendering.
\newblock In {\em CVPR}, 2020.

\bibitem{Kulon_2020_CVPR}
Dominik Kulon, Riza~Alp Guler, Iasonas Kokkinos, Michael~M. Bronstein, and
  Stefanos Zafeiriou.
\newblock Weakly-supervised mesh-convolutional hand reconstruction in the wild.
\newblock In {\em CVPR}, 2020.

\bibitem{Kulon2019}
Dominik Kulon, Haoyang Wang, Riza~Alp G{\"u}ler, Michael Bronstein, and
  Stefanos Zafeiriou.
\newblock Single image 3d hand reconstruction with mesh convolutions.
\newblock {\em BMVC}, 2019.

\bibitem{kundu20183d}
Abhijit Kundu, Yin Li, and James~M Rehg.
\newblock 3d-rcnn: Instance-level 3d object reconstruction via
  render-and-compare.
\newblock In {\em CVPR}, 2018.

\bibitem{kuo2020mask2cad}
Weicheng Kuo, Anelia Angelova, Tsung-Yi Lin, and Angela Dai.
\newblock Mask2cad: 3d shape prediction by learning to segment and retrieve.
\newblock {\em ECCV}, 2020.

\bibitem{Li2016}
Ke Li and Jitendra Malik.
\newblock Amodal instance segmentation.
\newblock In {\em ECCV}, 2016.

\bibitem{lim2013parsing}
Joseph~J Lim, Hamed Pirsiavash, and Antonio Torralba.
\newblock Parsing ikea objects: Fine pose estimation.
\newblock In {\em ICCV}, 2013.

\bibitem{Lin2014}
Tsung-Yi Lin, Michael Maire, Serge Belongie, James Hays, Pietro Perona, Deva
  Ramanan, Piotr Doll{\'a}r, and C~Lawrence Zitnick.
\newblock Microsoft coco: Common objects in context.
\newblock In {\em ECCV}, 2014.

\bibitem{Mandikal2020}
Priyanka Mandikal and Kristen Grauman.
\newblock Dexterous robotic grasping with object-centric visual affordances.
\newblock {\em arXiv:2009.01439}, 2020.

\bibitem{Meltzoff1995}
Andrew~N Meltzoff.
\newblock Understanding the intentions of others: re-enactment of intended acts
  by 18-month-old children.
\newblock {\em Developmental psychology}, 1995.

\bibitem{Michel2017}
Frank Michel, Alexander Kirillov, Eric Brachmann, Alexander Krull, Stefan
  Gumhold, Bogdan Savchynskyy, and Carsten Rother.
\newblock Global hypothesis generation for 6d object pose estimation.
\newblock In {\em CVPR}, 2017.

\bibitem{moon2018v2v}
Gyeongsik Moon, Ju Yong~Chang, and Kyoung Mu~Lee.
\newblock V2v-posenet: Voxel-to-voxel prediction network for accurate 3d hand
  and human pose estimation from a single depth map.
\newblock In {\em CVPR}, 2018.

\bibitem{mueller2018ganerated}
Franziska Mueller, Florian Bernard, Oleksandr Sotnychenko, Dushyant Mehta,
  Srinath Sridhar, Dan Casas, and Christian Theobalt.
\newblock Ganerated hands for real-time 3d hand tracking from monocular rgb.
\newblock In {\em CVPR}, 2018.

\bibitem{oikonomidis2011full}
Iason Oikonomidis, Nikolaos Kyriazis, and Antonis~A Argyros.
\newblock Full dof tracking of a hand interacting with an object by modeling
  occlusions and physical constraints.
\newblock In {\em ICCV}, 2011.

\bibitem{SMPL-X:2019}
Georgios Pavlakos, Vasileios Choutas, Nima Ghorbani, Timo Bolkart, Ahmed A.~A.
  Osman, Dimitrios Tzionas, and Michael~J. Black.
\newblock Expressive body capture: 3d hands, face, and body from a single
  image.
\newblock In {\em CVPR}, 2019.

\bibitem{2018-TOG-SFV}
Xue~Bin Peng, Angjoo Kanazawa, Jitendra Malik, Pieter Abbeel, and Sergey
  Levine.
\newblock Sfv: Reinforcement learning of physical skills from videos.
\newblock {\em ACM Trans. Graph.}, 2018.

\bibitem{Radosavovic2021}
Ilija Radosavovic, Xiaolong Wang, Lerrel Pinto, and Jitendra Malik.
\newblock State-only imitation learning for dexterous manipulation.
\newblock In {\em IROS}, 2021.

\bibitem{Ranftl2020}
Ren\'{e} Ranftl, Katrin Lasinger, David Hafner, Konrad Schindler, and Vladlen
  Koltun.
\newblock Towards robust monocular depth estimation: Mixing datasets for
  zero-shot cross-dataset transfer.
\newblock {\em TPAMI}, 2020.

\bibitem{romero2010hands}
Javier Romero, Hedvig Kjellstr{\"o}m, and Danica Kragic.
\newblock Hands in action: real-time 3d reconstruction of hands in interaction
  with objects.
\newblock In {\em ICRA}, 2010.

\bibitem{romero2017embodied}
Javier Romero, Dimitrios Tzionas, and Michael~J Black.
\newblock Embodied hands: Modeling and capturing hands and bodies together.
\newblock {\em TOG}, 2017.

\bibitem{rong2020frankmocap}
Yu Rong, Takaaki Shiratori, and Hanbyul Joo.
\newblock Frankmocap: Fast monocular 3d hand and body motion capture by
  regression and integration.
\newblock {\em arXiv:2008.08324}, 2020.

\bibitem{Sahasrabudhe2019}
Mihir Sahasrabudhe, Zhixin Shu, Edward Bartrum, Riza Alp~Guler, Dimitris
  Samaras, and Iasonas Kokkinos.
\newblock Lifting autoencoders: Unsupervised learning of a fully-disentangled
  3d morphable model using deep non-rigid structure from motion.
\newblock In {\em ICCV}, 2019.

\bibitem{Shan2020}
Dandan Shan, Jiaqi Geng, Michelle Shu, and David Fouhey.
\newblock Understanding human hands in contact at internet scale.
\newblock In {\em CVPR}, 2020.

\bibitem{sharp2015accurate}
Toby Sharp, Cem Keskin, Duncan Robertson, Jonathan Taylor, Jamie Shotton, David
  Kim, Christoph Rhemann, Ido Leichter, Alon Vinnikov, Yichen Wei, et~al.
\newblock Accurate, robust, and flexible real-time hand tracking.
\newblock In {\em CHI}, 2015.

\bibitem{simon2017hand}
Tomas Simon, Hanbyul Joo, Iain Matthews, and Yaser Sheikh.
\newblock Hand keypoint detection in single images using multiview
  bootstrapping.
\newblock In {\em CVPR}, 2017.

\bibitem{sridhar2016real}
Srinath Sridhar, Franziska Mueller, Michael Zollh{\"o}fer, Dan Casas, Antti
  Oulasvirta, and Christian Theobalt.
\newblock Real-time joint tracking of a hand manipulating an object from rgb-d
  input.
\newblock In {\em ECCV}, 2016.

\bibitem{sridhar2013interactive}
Srinath Sridhar, Antti Oulasvirta, and Christian Theobalt.
\newblock Interactive markerless articulated hand motion tracking using rgb and
  depth data.
\newblock In {\em ICCV}, 2013.

\bibitem{Sun2018}
Xingyuan Sun, Jiajun Wu, Xiuming Zhang, Zhoutong Zhang, Chengkai Zhang, Tianfan
  Xue, Joshua~B Tenenbaum, and William~T Freeman.
\newblock Pix3d: Dataset and methods for single-image 3d shape modeling.
\newblock In {\em CVPR}, 2018.

\bibitem{tagliasacchi2015robust}
Andrea Tagliasacchi, Matthias Schr{\"o}der, Anastasia Tkach, Sofien Bouaziz,
  Mario Botsch, and Mark Pauly.
\newblock Robust articulated-icp for real-time hand tracking.
\newblock In {\em Computer Graphics Forum}, 2015.

\bibitem{GRAB:2020}
Omid Taheri, Nima Ghorbani, Michael~J. Black, and Dimitrios Tzionas.
\newblock {GRAB}: A dataset of whole-body human grasping of objects.
\newblock In {\em ECCV}, 2020.

\bibitem{tekin2019h}
Bugra Tekin, Federica Bogo, and Marc Pollefeys.
\newblock H+o: Unified egocentric recognition of 3d hand-object poses and
  interactions.
\newblock In {\em CVPR}, 2019.

\bibitem{Tenenbaum2000}
Joshua~B Tenenbaum, Vin De~Silva, and John~C Langford.
\newblock A global geometric framework for nonlinear dimensionality reduction.
\newblock {\em Science}, 2000.

\bibitem{factored3dTulsiani17}
Shubham Tulsiani, Saurabh Gupta, David Fouhey, Alexei~A. Efros, and Jitendra
  Malik.
\newblock Factoring shape, pose, and layout from the 2d image of a 3d scene.
\newblock In {\em CVPR}, 2018.

\bibitem{tzionas2016capturing}
Dimitrios Tzionas, Luca Ballan, Abhilash Srikantha, Pablo Aponte, Marc
  Pollefeys, and Juergen Gall.
\newblock Capturing hands in action using discriminative salient points and
  physics simulation.
\newblock {\em IJCV}, 2016.

\bibitem{xiang2019monocular}
Donglai Xiang, Hanbyul Joo, and Yaser Sheikh.
\newblock Monocular total capture: Posing face, body, and hands in the wild.
\newblock In {\em CVPR}, 2019.

\bibitem{Xiang2018}
Yu Xiang, Tanner Schmidt, Venkatraman Narayanan, and Dieter Fox.
\newblock Posecnn: A convolutional neural network for 6d object pose estimation
  in cluttered scenes.
\newblock In {\em RSS}, 2018.

\bibitem{yang2019disentangling}
Linlin Yang and Angela Yao.
\newblock Disentangling latent hands for image synthesis and pose estimation.
\newblock In {\em CVPR}, 2019.

\bibitem{ye2016spatial}
Qi Ye, Shanxin Yuan, and Tae-Kyun Kim.
\newblock Spatial attention deep net with partial pso for hierarchical hybrid
  hand pose estimation.
\newblock In {\em ECCV}, 2016.

\bibitem{yuan2018depth}
Shanxin Yuan, Guillermo Garcia-Hernando, Bj{\"o}rn Stenger, Gyeongsik Moon, Ju
  Yong~Chang, Kyoung Mu~Lee, Pavlo Molchanov, Jan Kautz, Sina Honari, Liuhao
  Ge, et~al.
\newblock Depth-based 3d hand pose estimation: From current achievements to
  future goals.
\newblock In {\em CVPR}, 2018.

\bibitem{zhang2020phosa}
Jason~Y. Zhang, Sam Pepose, Hanbyul Joo, Deva Ramanan, Jitendra Malik, and
  Angjoo Kanazawa.
\newblock Perceiving 3d human-object spatial arrangements from a single image
  in the wild.
\newblock In {\em ECCV}, 2020.

\bibitem{zhou2020monocular}
Yuxiao Zhou, Marc Habermann, Weipeng Xu, Ikhsanul Habibie, Christian Theobalt,
  and Feng Xu.
\newblock Monocular real-time hand shape and motion capture using multi-modal
  data.
\newblock In {\em CVPR}, 2020.

\bibitem{Zhu2017}
Yan Zhu, Yuandong Tian, Dimitris Metaxas, and Piotr Doll{\'a}r.
\newblock Semantic amodal segmentation.
\newblock In {\em CVPR}, 2017.

\bibitem{zimmermann2017learning}
Christian Zimmermann and Thomas Brox.
\newblock Learning to estimate 3d hand pose from single rgb images.
\newblock In {\em ICCV}, 2017.

\bibitem{zimmermann2019freihand}
Christian Zimmermann, Duygu Ceylan, Jimei Yang, Bryan Russell, Max Argus, and
  Thomas Brox.
\newblock Freihand: A dataset for markerless capture of hand pose and shape
  from single rgb images.
\newblock In {\em ICCV}, 2019.

\end{thebibliography}
